\pdfoutput=1
\documentclass[11pt]{article}

\usepackage{times}
\usepackage{latexsym}

\usepackage[utf8]{inputenc}

\usepackage{microtype}

\usepackage{inconsolata}

\usepackage{multirow}
\usepackage{bbm}
\usepackage{amsmath}
\usepackage{graphicx}
\usepackage{subcaption}
\usepackage[numbers]{natbib}
\usepackage{hyperref}

\title{Discriminative Language Model as Semantic Consistency Scorer for Prompt-based Few-Shot Text Classification}

\author{Zhipeng Xie and Yahe Li\\
School of Computer Science \\
  Fudan University, Shanghai, China \\
\texttt{xiezp@fudan.edu.cn} \\}

\begin{document}

\maketitle

\begin{abstract}
This paper proposes a novel prompt-based finetuning method (called DLM-SCS) for few-shot text classification by utilizing the discriminative language model ELECTRA that is pretrained to distinguish whether a token is original or generated. The underlying idea is that the prompt instantiated with the true label should have higher semantic consistency score than other prompts with false labels. Since a prompt usually consists of several components (or parts), its semantic consistency can be decomposed accordingly. The semantic consistency of each component is then computed by making use of the pretrained ELECTRA model, without introducing extra parameters. Extensive experiments have shown that our model outperforms several state-of-the-art prompt-based few-shot methods.
\end{abstract}

\section{Introduction}

Nowadays, with the upsurge of interest in a wide range of pretrained language models, the \textit{pretraining-finetuning} paradigm~\cite{radford2018,dong2019} has become a de facto standard for various downstream NLU and NLG tasks. Different language models usually have different scopes of application.
%Pretrained language models (PLM) 
\textit{Auto-regressive language models} (ARLM) such as GPT-3~\cite{brown2020} and Ernie-3~\cite{sun2021} predict the next token based on all the previous ones, usually in the left-to-right order. Since it is trained to encode a uni-directional context, it is not effective at downstream NLU tasks that often require bidirectional context information. In addition, these models are large and costly to finetune, or even not available publicly, which makes them impossible to use in the pretraining-finetuning paradigm. \textit{Masked language models} (MLM) such as BERT~\cite{devlin2019} and RoBERTa~\cite{liu2019} mask some tokens in inputs and are trained to reconstruct the original tokens based on their bidirectional surrounding context, which is often preferable in NLU tasks such as text classification but not applicable in NLG.

\textit{Conventional finetuning method} for downstream text classification task usually builds up a classification head together with additional parameters on top of the special \verb|[CLS]| token from scratch and fine-tunes the whole model. Such models work well with abundant training examples in rich data regimes, but will be cornered in the few-shot scenario, not to mention the zero-shot, because of the gap between the pretraining and the downstream tasks.

%To bridge the gap, In prompt-based finetuning, downstream tasks are reformulated to look more like the pretraining tasks solved during the original LM training with the help of a textual prompt.
%Prompt-based finetuning methods have exhibited promising few-shot performance. 

\begin{figure*}
\centering
\includegraphics[scale=0.23]{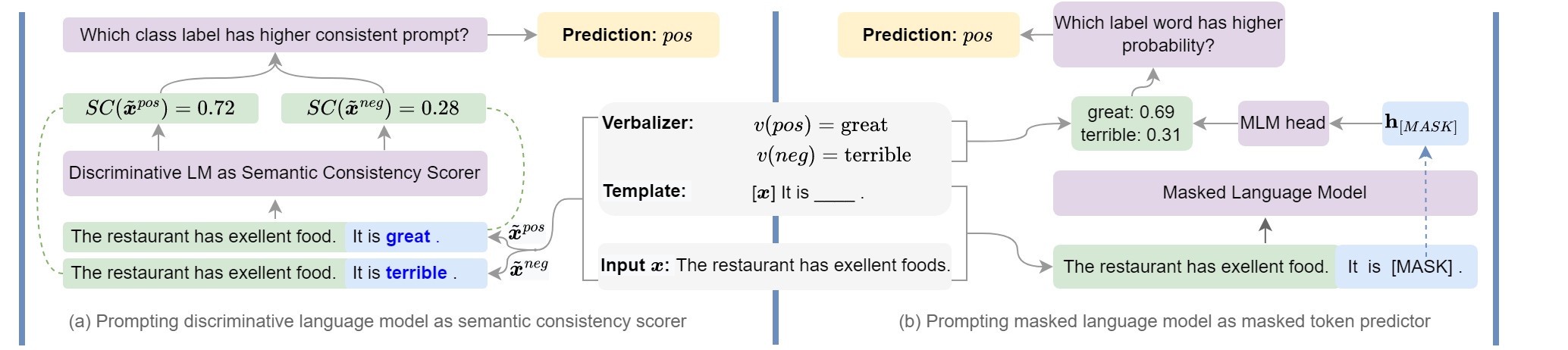}
\caption{A schematic illustration of (a) our proposed DLM-SCS (Discriminative Language Model as Semantic Consistency Scorer), comparing to that of (b) traditional prompt-based model that uses masked language model as masked token predictor.}
\label{fig:framework}
\end{figure*}

Initiated by the in-context learning of the GPT series~\cite{radford2018,radford2019,brown2020}, prompt-based method was first developed for zero-shot learning, and then studied by PET and iPET~\cite{schick2021} for finetuning. After that, prompt-based learning methods have become increasingly popular, and have been proven to work effectively under few-shot or even zero-shot setting. To bridge the gap between the downstream task and the pretrained task, these methods transform downstream tasks into the same (or similar) form as the pretraining tasks solved during the original LM training with the help of textual prompts. Most existing prompt-based methods are using generative prompts that contain answer slots for various pretrained language models to fill in~\cite{liu2021}. As to the downstream text classification tasks, most works have been directed against pretrained masked language models (MLMs) by formulating downstream tasks as a masked language modeling task~\cite{schick2021,schick2021b,gao2021}. A \textit{template} converts the original input example $\boldsymbol{x}_{\text{in}}$ into a textual string (called \textit{prompt}) $\tilde{\boldsymbol{x}}$ that contains an unfilled \verb|[MASK]| slot. A \textit{verbalizer} is used to represent each class with a \textit{label word} from the vocabulary. The model makes the prediction according to the probabilities of filling the \verb|[MASK]| token with the label words. Such a prompt is called a \textit{generative prompt}, which usually contains an an unfilled \verb|[MASK]| as the answer slot, and the pretrained masked language model is finetuned to generate a correct label name to fill this answer slot. A simple prompt-based framework that treats MLM as masked token predictor for text classification is illustrated in Figure~\ref{fig:framework}(b).  

Until the very recent, two prompt-based finetuning methods~\cite{yao2022,xia2022} have been proposed to exploit the pretrained ELECTRA~\cite{clark2020} which is a \textit{discriminative language model} (DLM). In contrast to the generative prompts, they use the prompts that contains no answer slot, which we call ``\textit{discriminative prompts}'' and can be seen as the unmasked prompts which use label word(s) to fill in the \verb|[MASK]| of generative prompts. The pretrained ELECTRA model is then applied on these discriminative prompts and tells us which label word is the original token (i.e., \textit{not} a replaced token). 
However, these methods confine themselves only on the label word(s) in the discriminative prompts and expect the discriminative model to identify the semantic inconsistency incurred by the incorrect label words. This limited evidence is far from what can be obtained from the discriminative language model, and some available evidence is missing (Please refer to Section~\ref{sec:motivation} for a simple motivating example). 

The work done in this paper follows the thread of prompting the discriminative language model for few-shot text classification. The basic idea is that the DLM head can detect the discrepancy between inputs and label words. On the one hand, given an input example (a sentence or a sentence pair) and its true label, the DLM head is expected to assign low scores (or logits) to the salient tokens in the input example and the true label word. If a false label is given, it is desirable that the DLM head will assign high scores to both the false label word and the salient tokens in the input example.

%The food tastes 
%The standard fine-tuning approach for text classification is 
%Prompt-based approaches convert ... 

To squeeze the most out of a pretrained language model such that it works best on a downstream few-shot learning task, three prerequisites are considered in designing a prompt-based method of finetuning a pretrained discriminative language model: 
\begin{itemize}
\item \textbf{Prerequisite 1 (Task Compatibility):} As stated by most prompt tuning methods, the downstream task should be transformed into the same (or highly similar) form as the pretraining task, such that no (or few) additional parameters need introduced. 
\item \textbf{Prerequisite 2 (Input Compatibility):} The prompt template should be in the same form as the training data of the pretrained language model, such that the discriminative prompts are as much natural as possible. As a consequence, the pretrained language model can process them well and easily, without having to be tuned too far away.
\item \textbf{Prerequisite 3 (Evidence/Information Abundance):} Last but not the least, the method should try its best to obtain and aggregate as much evidence and/or information as possible for decision making. Due to the nature of few-shot learning, it is unstable and has a big variance, and thus the aggregation of more evidence would be helpful in reducing the variance. 
\end{itemize}

%Prompt-based learning turns the objective into a softmax distribution over all verbalizers of a prompt template~\cite{clark2020}.

The contribution of this paper is threefold: (1) We propose a novel framework DLM-SCS\footnote{We shall publicize all the source code and related resources once the paper gets accepted or published.} for few-shot text classification, which uses the pretrained discriminative language model ELECTRA as the semantic consistency scorer. (2) We design a method to measure the semantic consistency of a subsequence in the input prompt on the basis of the discriminative head of ELECTRA which can only measure the semantic inconsistency of each single token, and then use it to instantiate the framework into a concrete prompt-based finetuning model (also called DLM-SCS). (3) The proposed method has achieved the state-of-the-art performance on a variety of downstream sentence classification and sentence-pair classification tasks.

\section{Related Work}

\subsection{Prompting MLM for Text Classification}

Existing prompt-based learning methods for text classification usually reformulate the downstream text classification task into a cloze question task, and then finetune a pretrained masked language model to generate the most likely label word in the unfilled \verb|[MASK]| position of the generative prompt~\cite{schick2021}. A lot of research effort has been devoted to the automatic construction of prompt templates and label words. \citet{schick2020} and \citet{schick2021} studied the automatic identification of label words. \citet{gao2021} made use of the pretrained seq2seq model T5~\cite{raffel2020} to generate template tokens in the template search process. Motivated by idea of in-context learning from GPT series~\cite{brown2020}, \citet{gao2021} used a single unmasked example prompt (called a \textit{demonstration}) as additional context, which can boost the performance of prompt-based few-shot text classification task. \citet{park2022} made two extensions by multiple demonstrations and soft demonstration memory. In addition, \citet{zhang2021} proposed the DART method that optimizes the differentiable prompt template and label words by error backpropagation.

%Standard fine-tuning method first appends a special token \verb|[CLS]| to the input $\mathbf{x}$, then employs PLM to encode it into a sequence of hidden representations, and finally uses a classifier ()
%Given a masked language model, a prompt usually consists of a template function that converts the input to a prompt input, and a verbalizer that maps class labels to label words. 

%\citet{schick2021} proposed 

\subsection{Prompting DLM for Text Classification}

Two prompt-based text classification methods for finetuning discriminative language model (DLM) have been recently proposed~\cite{yao2022,xia2022}. These methods reformulate text classification task into a discriminative language modeling problem, and predict the class label of an input example by using the DLM head to identify which label name is the original token instead of a replaced one. The DPT method proposed in \cite{yao2022} fills the input text $\boldsymbol{x}$ into the following template of discriminative prompt: 
$$\verb|[CLS] | \boldsymbol{x}\verb| Class: |v(l_1), v(l_2), \dots, v(l_n).\verb| [SEP]|$$
where the verbalizer $v(\cdot)$ maps each class label $l_i$ ($1\le i\le n$) to a distinct label word.
Then the DLM head is used to judge which label word is proper in the context. It should be noted that DPT is not designed and also not suited for few-shot learning, because it does not satisfy the \textbf{Prerequisite 2} of \textit{input compatibility}. As shown in Section~\ref{sec:experiments}, DPT cannot work well in few-shot scenario.
%The label with the highest probability of being an original token ...
%When we input a prompt of an input and its false label to a DLM, it is expected that the label name
%\cite{yao2022} 
%\cite{xia2022}

The other method, PromptELECTRA~\cite{xia2022}, was designed for few-shot text classification. Given an input example $\boldsymbol{x}$, it will generate one discriminative prompt for each possible class label. Thus, there are $n$ discriminative prompts for $\boldsymbol{x}$. The DLM head is used to output the label word that has the highest probability of being original token in its corresponding prompt. This method satisfies \textbf{Prerequisite 1} and \textbf{2}, but it is not enough with respect to \textbf{Prerequisite 3} because it makes decision based on the only evidence from the candidate label words.
%In the prompt-based fine-tuning of a masked language model (MLM), each example of the downstream task is converted into a natural language prompt for the model to auto-complete. For example, in a binary sentiment classification task such as SST-2, given an input sentence $\boldsymbol{x}$ of ``\textit{The restaurant has excellent foods.}'', the prompt for the MLM model to fill in the 

\section{Background and Motivation}

\subsection{The Pretrained ELECTRA Language Model}
As a pretrained discriminative language model, ELECTRA~\cite{clark2020} consists of an encoder and a discriminative head. The encoder first maps a sequence of input tokens $\boldsymbol{x}=\left[x_1,x_2,\dots,x_n\right]$ into a sequence of contextualized vector representations $\left[\mathbf{h}_1,\mathbf{h}_2,\dots,\mathbf{h}_n\right]$, and the discriminative head then predicts whether each token $x_t$ $(1\le t\le n)$ is a ``real'' or ``replaced'' token. In particular, the discriminative head simply applies a linear layer with the sigmoid activation function on the contextualized representation $\mathbf{h}_t$ of $x_t$:
\begin{equation}
P_{\text{DLM}}(x_t, \boldsymbol{x})=\mathop{\text{sigmoid}}\left(\mathbf{w}^\top\mathbf{h}_t\right)
\end{equation}
where $\mathbf{w}$ denotes the parameter vector. In this paper, we interpret the value of $\mathbf{w}^\top\mathbf{h}_t$ as the \textit{unnormalized semantic inconsistency score} of token $x_t$ in the context $\boldsymbol{x}$. The larger the value of $\mathbf{w}^\top\mathbf{h}_t$ is, the more semantically inconsistent the token $x_t$ is, and the more likely the token $x_t$ is a replaced token in $\boldsymbol{x}$. 

%, a pretraining task called \textit{replaced token detection} was introduced to distinguish real input tokens from plausible but synthetically generated replacements. That is, for the $i$-th token $x_i$ in an input $\mathbf{x}$, a discriminative head is used to tell whether (or how likely) it is a real or an artificial token based on its hidden representation $\mathbf{h}_i$.  
%The ELECTRA model jointly trains a discriminator and a smaller generator 
%\begin{equation}
%-\sum_{i}\left(\mathbbm{1}\right)%(x_i^\'=x_i)\log \mathcal(H)(c(x_i)) + \mathbff(x_i^'\neq x_i)\log(1-\mathcal(H)(c(x_i^'))\right)
%\end{equation}

\subsection{Motivation}
\label{sec:motivation}

\begin{figure}
\centering
\includegraphics[scale=0.35]{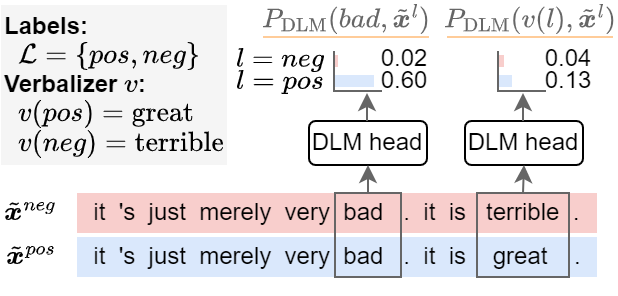}
\caption{A simple example that motivates the DLM-SCS model.}
\label{fig:example}
\end{figure}

Figure~\ref{fig:example} shows a simple input sentence from the sentiment classification task SST-2:
$$\boldsymbol{x}=\text{``it's just merely very bad.''}$$
The verbalizer $v$ maps the class label $pos$ (positive) to the label word ``great'', and the class label $neg$ (negative) to ``terrible''. Therefore, two discriminative prompts ($\tilde{\boldsymbol{x}}^{neg}$ and $\tilde{\boldsymbol{x}}^{pos}$) are generated by using the template ``$\boldsymbol{x}$ It is $v(l)$'', one for each class label $l\in\mathcal{L}=\{pos,neg\}$.

Applying the discriminative head of the pretrained ELECTRA-large model to the label words: ``terrible'' in $\tilde{\boldsymbol{x}}^{neg}$ and ``great'' in $\tilde{\boldsymbol{x}}^{pos}$, we observe that $P_{\text{DLM}}(\text{terrible}, \tilde{\boldsymbol{x}}^{neg})=0.04$ and $P_{\text{DLM}}(\text{great}, \tilde{\boldsymbol{x}}^{pos})=0.13$. This evidence thus supports the conclusion that the input sentence is more likely to be negative, because the discriminative prompt $\tilde{\boldsymbol{x}}^{neg}$ of $neg$  is semantically more consistent with respect to its label word (or in other words, $v(neg)=\text{``terrible''}$ is less likely to be a replaced token).

Besides the label tokens, the discriminative prompts contain tokens from the original input sentence, which may also provide us some evidence about the classification decision. As illustrated in Figure~\ref{fig:example}, by applying the DLM head on the token ``bad'' in the discriminative prompts, we get $P_{\text{DLM}}(\text{bad}, \tilde{\boldsymbol{x}}^{neg})=0.02$ and $P_{\text{DLM}}(\text{bad}, \tilde{\boldsymbol{x}}^{pos})=0.60$, which also support the same conclusion that the input sentence is negative. This evidence from the token ``bad'' is even stronger than the evidence provided by the label words.

\section{Method}

This section is devoted to a prompt-based framework for finetuning discriminative language models.
The main thrust is to treat prompt-based text classification as a task of semantic consistency scoring, and calculate the semantic consistency of a prompt as a weighted average of the semantic consistency scores of multiple components (or parts) in the prompt.  

Let $\mathcal{L}$ be the set of class labels for the target text classification task at hand. A verbalizer $v$ is an injective function that maps each class label to a single token from $M$'s vocabulary, $v:\mathcal{L}\to V$. 
We simply adopt the manual prompt templates used in the previous work~\cite{gao2021}. 
For sentence classification task, given an input example of single sentence $\boldsymbol{x}_\text{in}=\boldsymbol{x}^{(1)}$, we can generate a discriminative prompt $\tilde{\boldsymbol{x}}^l$ for each label $l\in\mathcal{L}$:
\begin{equation}
\tilde{\boldsymbol{x}}^l=\verb|[CLS] | \boldsymbol{x}^{(1)} \verb| It is | v(l)\verb| . | \verb|[SEP]|
\end{equation}
For sentence-pair classification task, given an input example of sentence pair $\boldsymbol{x}_{\text{in}}=\left(\boldsymbol{x}^{(1)}, \boldsymbol{x}^{(2)} \right)$ and a label $l\in\mathcal{L}$, we can generate the discriminative prompt $\tilde{\boldsymbol{x}}^l$ as:
\begin{equation}
\tilde{\boldsymbol{x}}^l=\verb|[CLS] | \boldsymbol{x}^{(1)} \verb| ? | v(l)\verb| , | \boldsymbol{x}^{(2)}\verb| [SEP]|
\end{equation}
Therefore, there are $|\mathcal{L}|$ discriminative prompts $\{\tilde{\boldsymbol{x}}^l|l\in\mathcal{L}\}$ (one for each class label) generated for each input example. 

The decision criterion of DLM-SCS model is based on the assumption that the discriminative prompt of true class label (\textit{true prompt} in short) is semantically more consistent than the other discriminative prompts of false labels (\textit{false prompts}).
The class label $l$ whose prompt $\tilde{\boldsymbol{x}}^l$ has the highest semantic consistency is chosen as the predicted label
\begin{equation}
\hat{l} = \mathop{\arg\max}_{l\in\mathcal{L}} SC(\tilde{\boldsymbol{x}}^l)
\end{equation}
where $SC(\tilde{\boldsymbol{x}}^l)$ denotes the semantic consistency of the discriminative prompt $\tilde{\boldsymbol{x}}^l$. Figure~\ref{fig:framework}(a) demonstrates this idea of using DLM as a semantic consistency scorer for text classification task.

Next, we move to the problem of how to calculate the semantic consistency $SC(\tilde{\boldsymbol{x}}^l)$ of a discriminative prompt $\tilde{\boldsymbol{x}}^l$. Since each prompt consists of several components (or parts), its semantic consistency can be decomposed accordingly. In particular, the semantic consistency of $\tilde{\boldsymbol{x}}^l$ is calculated as a weighted average of the semantic consistencies of some parts in $\tilde{\boldsymbol{x}}^l$:
\begin{multline}
%\scriptstyle
SC(\tilde{\boldsymbol{x}}^l) = \lambda_0\cdot sc\left(v(l), \tilde{\boldsymbol{x}}^l\right)\\+\sum_{\boldsymbol{x}^{(i)}\in\boldsymbol{x}_{\text{in}}}\lambda_i\cdot sc\left(\boldsymbol{x}^{(i)}, \tilde{\boldsymbol{x}}^l\right)
\label{eq:SC}
\end{multline}
where:
\begin{itemize}
\item The term of form $sc(\boldsymbol{s},\tilde{\boldsymbol{x}}^l)$ denotes the semantic consistency of a token subsequence $\boldsymbol{s}$ in the context of discriminative prompt $\tilde{\boldsymbol{x}}^l$. Here, $v(l)$ is  seen as a subsequence of single token.
\item The $\lambda_i$'s are hyperparameters indicating the relative importance of the prompt parts ($i\in\{0,1\}$ for sentence classification task, while $i\in\{0,1,2\}$ for sentence-pair classification).
\end{itemize}

Given a token subsequence $\boldsymbol{s}$ in the discriminative prompt $\tilde{\boldsymbol{x}}^l$, its semantic consistency can be simply measured by the softmax activation function over the mean negative logits of its tokens in the $|\mathcal{L}|$ discriminative prompts:
\begin{equation}
%sc(\boldsymbol{s},\boldsymbol{x}_{\text{in}}^l) = 1 - \sigma\left( \sum_{x\in\boldsymbol{s}}\mathbf{w}^\top\mathbf{h}_x^l\right)
sc(\boldsymbol{s},\tilde{\boldsymbol{x}}^l) = \frac{\exp\left(-\frac{1}{|\boldsymbol{s}|}\sum_{x\in\boldsymbol{s}}\mathbf{w}^\top\mathbf{h}_x^l\right)}{\sum_{l'\in\mathcal{L}}\exp\left(-\frac{1}{|\boldsymbol{s}|}\sum_{x\in\boldsymbol{s}}\mathbf{w}^\top\mathbf{h}_x^{l'}\right)}
\label{eq:unisc}
\end{equation}
where $\mathbf{h}_x^l$ denotes the contextualized representation of token $x$ in the discriminative prompt $\tilde{\boldsymbol{x}}^l$.
Here,  because the value of $\mathbf{w}^\top\mathbf{h}_x^l$ denotes the semantic inconsistency score of $x$ (i.e., the logit that $x$ is a replaced token), the minus sign is introduced into the exponent part in order to transform it into the score of semantic consistency. %where $\sigma(\cdot)$ denotes the sigmoid activation function.  
As to the semantic consistency of the label word $v(l)$ in $\tilde{\boldsymbol{x}}^l$, it is calculated with the following equation:
\begin{equation}
%sc(\boldsymbol{s},\boldsymbol{x}_{\text{in}}^l) = 1 - \sigma\left( \sum_{x\in\boldsymbol{s}}\mathbf{w}^\top\mathbf{h}_x^l\right)
sc(v(l),\tilde{\boldsymbol{x}}^l) = \frac{\exp\left(-\mathbf{w}^\top\mathbf{h}_{v(l)}^l\right)}{\sum_{l'\in\mathcal{L}}\exp\left(-\mathbf{w}^\top\mathbf{h}_{v(l')}^{l'}\right)}
\label{eq:unisclabel}
\end{equation}

In Equation~\ref{eq:unisc}, all tokens in $\boldsymbol{s}$ are treated equally and their negative logits are simply averaged. However, different tokens should be of different importance with respect to the semantic consistency. Therefore, we make slight modification to Equation~\ref{eq:unisc} by weighting each token with its inverse document frequency (IDF), as below:
\begin{equation}
%sc(\boldsymbol{s},\boldsymbol{x}_{\text{in}}^l) = 1 - \sigma\left( \sum_{x\in\boldsymbol{s}}\mathbf{w}^\top\mathbf{h}_x^l\right)
sc(\boldsymbol{s},\tilde{\boldsymbol{x}}^l) = \frac{\exp\left(-\frac{\sum_{x\in\boldsymbol{s}}\text{idf}(x)\mathbf{w}^\top\mathbf{h}_x^l}{\sum_{x\in\boldsymbol{s}}\text{idf}(x)}\right)}{\sum_{l'\in\mathcal{L}}\exp\left(-\frac{\sum_{x\in\boldsymbol{s}}\text{idf}(x)\mathbf{w}^\top\mathbf{h}_x^{l'}}{\sum_{x\in\boldsymbol{s}}\text{idf}(x)}\right)}
\label{eq:idfsc}
\end{equation}
where $\text{idf}(x)$ of token $x$ is set to be the inverse document frequency of the word that $x$ belongs to. The inverse document frequency of words can be easily calculated from a large-scale unlabeled text corpus. In this paper, we simple obtain the IDF values from the sentences in the full training set, and normalize the IDF values into the range of $[0,1]$.

%The semantic consistency of a prompt can be measured by combining the semantic consistency of salient tokens in the prompt.
%In turn, the semantic consistency of a prompt part 

%Therefore, there are two main components in the process. One is how to calculate the semantic consistency of a token in the prompt, the second is which tokens are salient in the prompt, and the other is how to combine the semantic consistencies of these salient tokens. 

%Our method works on the basis of an underlying assumption: for a given prompt of the input with a false class label, at least one token in the input will be identified as replaced token by the DLM. 
%The token of false class label is semantically inconsistent in the false prompt and will be detected as a replaced token by the DLM.

\subsection{The Loss Function}

Given a training example $\boldsymbol{x}_\text{in}$, each discriminative prompt $\tilde{\boldsymbol{x}}^{l}$ ($l\in\mathcal{L}$) contains $m$ components/parts, where $m=2$ for single-sentence classification, while $m=3$ for sentence-pair classification. Each component/part corresponds to a subsequence of token in the prompt. The semantic consistency scorer of each prompt part actually outputs a probability distribution over the label set $\mathcal{L}$. Therefore, we use the cross-entropy function as the loss of each prompt part.

For the prompt part of label words, its loss is:
\begin{equation}
loss_0 = - \log sc(v(l^*), \tilde{\boldsymbol{x}}^{l^*})
\end{equation}
where $l^*$ is the true class label of the input example $\boldsymbol{x}_\text{in}$.

For the part of a sentence $\boldsymbol{x}^{(i)}$ in the input example ($i=1$ for single-sentence classification, while $i\in\{1,2\}$ for sentence-pair classification), the loss is measured as:
%The inconsistency of the first sentence with respect to the second sentence and the class label is measured as:
\begin{equation}
loss_{i} = - \log sc(\boldsymbol{x}^{(i)}, \tilde{\boldsymbol{x}}^{l^*})
\end{equation}
%where $\mathbf{h}_x^l$ denotes the hidden representation of token $x$ in the $l$-label prompt.

%with true class label $l$, the semantic consistency of label names $sc(v(l'),)
%Since the output semantic consistency values is a probability distribution over the class labels, we simply adopt the plain cross-entropy function as the loss:
%\begin{equation}
%L = \sum_{l\in\mathcal{L}} \log SC(\tilde{\boldsymbol{x}}_{\text{in}}^l)
%\end{equation}

Therefore, the total loss of the training example $\boldsymbol{x}_\text{in}$ is defined as the weighted average of the losses of its parts:
\begin{equation}
Loss = \sum_{0\le i< m} \lambda_i\cdot loss_i
\end{equation}
where the hyperparameters $\lambda_i$'s are the same as the ones used in Equation~\ref{eq:SC}.

\subsection{Model Optimization}
\label{sec:optimization}

For model training, we adopt AdamW algorithm and set a linear learning rate variation with warmup ratio of 0.05. For all the datasets, we take learning rate as 1e-5, and batch size as 2 input examples. For each trial, we train the model for 15 epochs, validate the performance every 50 steps, and take the best checkpoint. Early stopping is used to avoid overfitting. 

As to the hyperparameters $\lambda_i$'s, we set $\lambda_1=1-\lambda_0$ for single-sentence classification, while $\lambda_1=\lambda_2=\frac{1-\lambda_0}{2}$ for sentence-pair classification. The best value of $\lambda_0$ is chosen based on its performance on the development set by a grid search from 0.0 to 1.0 with step size $\frac{1}{30}$. The values of $\lambda_0$ chosen for the experimental tasks are provided in Appendix~\ref{sec:hyperparameter}.

%For a given sentence classification task, in a true prompt, it is expected that 
%The basic idea of our prompting method is that the prompt with the true label is more consistent than the other false labels.
%The pretrained DLM
%\section{Key Problem}
%Given an input, its true label word is expected to be consistent with it, and false label word is expected to be inconsistent. Therefore, the loss is easily calculated as:
%\begin{equation}
%\end{equation}

%The main obstacle exists in that we do not know which tokens of the input are expected to be in...

%One extreme is to treat all tokens separately and equally. However, this will lead to many false supervision signals in the finetuning process.

%Given a training example $\mathbf{x}_{\text{in}}$, there is one true prompt and $|\mathcal{L}|-1$ false prompts.  

\section{Experimental Results}
\label{sec:experiments}

\begin{table*}
\centering
%\small
\scalebox{0.95}{
\begin{tabular}{lcccccccccc}
\hline\hline
\multirow{2}{*}{\textbf{Model}} & \textbf{SNLI} 	& \textbf{MNLI} & \textbf{QNLI} & \textbf{RTE} & \textbf{MRPC} & \textbf{QQP} & \textbf{SST-2} & \textbf{SST-5}  & \textbf{MR}  &  \textbf{CR} \\
                       & (acc)             & (acc)         &(acc)          & (acc)        & (F1)          & (F1)         & (acc)          & (acc)           & (acc)        & (acc)        \\                     
\hline\hline
Fine-tuning      & 48.4 & 45.8 & 60.2 & 54.4 & 76.6 & 60.7 & 81.4 & 43.9 & 76.9 & 75.8\\
\hline
LM-BFF (man) 	& 77.2 & 68.3 & 64.5 & 69.1 & 74.5 & 65.5 & 92.7 & 47.4 & 87.0 & 90.3\\
 +demonstrations& 79.7 & 70.7 & 69.2 & 68.7 & 77.8 & 69.8 & 92.6 & 50.6 & 86.6 & 90.2\\
\hline
LM-BFF (auto) 	& 77.1 & 68.3 & 68.3 & 73.9 & 76.2 & 67.0 & 92.3 & 49.2 & 85.5 & 89.0\\
+demonstrations & 77.5 & 70.0 & 68.5 & 71.1 & 78.1 & 67.7 & 93.0 & 49.5 & 87.7 & 91.0\\
\hline
%P-Tuning        & 72.3 & 61.5 & 64.3 & -    & 74.5 & 65.6 &&&&\\  
%\hline
DART            & 75.8 & 67.5 & 66.7 & 68.7 & \textbf{78.3} & 67.8 & 93.5 & 49.6 & 88.2 & \textbf{91.8} \\
\hline
DPT             & 47.4 & 39.0 & 54.6 & 50.2 & 76.4 & 56.1 & 92.6 & 44.0 & 89.5 & 91.2 \\
\hline
PromptELECTRA   & 79.1 & 65.8 & 70.9 & 68.2 & 73.5 & 63.1 & 93.1 & 51.4 & 89.4 & 90.2 \\
\hline\hline
DLM-SCS (ours)  & \textbf{82.2} & \textbf{71.0} & \textbf{77.0} & \textbf{75.0} & \textbf{78.3} & \textbf{72.2} & \textbf{93.6} & \textbf{51.5} & \textbf{90.2} & 91.0 \\\hline
\end{tabular}
}
\caption{Performance evaluation on 6 sentence-pair classification tasks and 4 sentence classification tasks. The reported performance metrics are averaged over the same set of 5 random seeds, each random seed is used to sample 16 training example per class for training set, and the development set is of the same size as the training set.}
\label{tab:comparison}
\end{table*}

\begin{table*}
\centering
%\small
\scalebox{0.95}{
\begin{tabular}{lcccccccccc}
\hline\hline
\textbf{Model}			& \textbf{SNLI} 	& \textbf{MNLI} & \textbf{QNLI} & \textbf{RTE} & \textbf{MRPC} & \textbf{QQP} & \textbf{SST-2} & \textbf{SST-5}  & \textbf{MR}  &  \textbf{CR} \\
\hline\hline
(full) DLM-SCS      & 82.2 & 71.0 & 77.0 & 75.0 & 78.3 & 72.2 & 93.6 & 51.5 & 90.2 & 91.0 \\
\hline
-w.o. token weight    & 78.2 & 70.0 & 73.4 & 73.6 & 76.9 & 69.6 & 93.0 & 48.8    & 90.3 & 90.3 \\
-only label word      & 76.5 & 64.6 & 69.0 & 71.8 & 74.8 & 64.2 & 93.7 & 51.1    & 88.8 & 90.4\\
\hline
\end{tabular}
}
\caption{Ablation study. The row ``w.o. token weight'' removes the IDF weights from Equation \ref{eq:idfsc} (i.e., use Equation~\ref{eq:unisc} instead of Equation~\ref{eq:idfsc}). The row ``only label word'' removes the semantic consistency of the input example, and use simply the label words to measure the semantic consistency of a discriminative prompt.}
\label{tab:ablation}
\end{table*}

\subsection{Experimental Setting}
To evaluate the performance of our approach, we follow the experimental setting from \cite{gao2021}. For each task, we take only $K$ training examples per class for the training set $\mathcal{D}_\text{train}$, and thus the total number of training examples is $K\times|\mathcal{L}|$. A development set $\mathcal{D}_\text{dev}$ of the same size as the few-shot training set is employed for model selection and hyper-parameter tuning. Unless specified otherwise, the value of $K$ is set to 16 by default, and the reported performance metrics are averaged over the same set of 5 random seeds. 

We conduct extensive experiments on a variety of NLP tasks and make comparison with several state-of-the-art few-shot learning methods. 
We evaluate our model on 4 sentence classification tasks (SST-2, SST-5, MR, and CR) and 6 sentence-pair classification tasks (SNLI, MNLI, QNLI, RTE, MRPC and QQP). The characteristics of these tasks are listed in Appendix~\ref{sec:appdatasets}, and the templates and label words used are provided in Appendix~\ref{sec:apptemplates}.

\subsection{Main Results}

To evaluate the performance of our models, we make a comparison with the conventional fine-tuning method and several state-of-the-art prompt-based finetuning methods:
\vspace{-2mm}
\begin{itemize}
\setlength\itemsep{0.1em}
\item \textbf{Fine-tuning}: The conventional fine-tuning of Roberta-Large in the few-shot experimental setting.
\item \textbf{LM-BFF(man)}: The better few-shot fine-tuning of language models with manual prompts~\cite{gao2021}. 
\item \textbf{LM-BFF(auto)}: The better few-shot fine-tuning of language models with automatically searched templates~\cite{gao2021}. For both LM-BFF (man) and LM-BFF (auto), ``+demonstrations'' means incorporating demonstrations as additional context, which leads to performance gains in majority of tasks as indicated in \cite{gao2021}.
\item \textbf{DART}: The differentiable prompt framework proposed in \cite{zhang2022}, where the prompt template and the target labels are differentially optimized with backpropagation.
\item \textbf{DPT}: The prompt tuning framework for discriminative PLMs proposed in \cite{yao2022}. Although DPT does not aim at few-shot learning, we also include it in our experimental comparison. Its original implementation\footnote{\url{ https://github.com/thunlp/DPT}} can only deal with sentence classification tasks, we make a straightforward extension by using the manual template: ``\verb|[CLS]| $\boldsymbol{x}^{(1)}$ $\boldsymbol{x}^{(2)}$ \verb|Class:| $v(l_1)$, $v(l_2)$, ..., $v(l_n)$ \verb|[SEP]|'' for a sentence-pair classification task with $n$ classes.
\item \textbf{PromptELECTRA}\footnote{code available at \url{https://github.com/facebookresearch/ELECTRA-Fewshot-Learning}}: The few-shot learning framework with discriminative pretrained models in \cite{xia2022}. 
\end{itemize}

Table~\ref{tab:comparison} reports the few-shot finetuning results of these methods on large-sized PLMs, where \textbf{DLM-SCS}, \textbf{PromptELECTRA} and \textbf{DPT} are based on the \textbf{ELECTRA-large} model, while all the other methods are based on the \textbf{Roberta-large} model. 
It can be easily observed that our \textbf{DLM-SCS} model is the best performer and has achieved the best performance on 9 of the 10 tasks among all the competitors. The \textbf{Fine-tuning} and \textbf{DPT} are not designed for the few-shot setting and have the worst performance. The conventional \textbf{Fine-tuning} method does not satisfy the \textbf{Prerequisite 1} of \textit{Task Compatibility} because the finetuning task does not match the pretraining task and additional parameters get introduced, while \textbf{DPT} does not satisfy the \textbf{Prerequisite 2} of \textit{Input Compatibility} because the prompts used are not natural. \textbf{DLM-SCS} outperforms \textbf{PromptELECTRA} on all the ten tasks, because \textbf{DLM-SCS} makes use of more evidence in the prompt than \textbf{PromptELECTRA}, which has manifested the value of \textbf{Prerequisite 3} (\textit{Evidence/Information Abundance}).

\begin{table}
\centering
%\small
\scalebox{0.95}{
\begin{tabular}{lcccccccccc}
\hline\hline
\textbf{Model}			& \textbf{MM-SP} & \textbf{MM-S} \\
\hline\hline
Fine-tuning     & 57.7 & 69.5 \\
\hline
LM-BFF (man) 	& 69.9 & 79.4 \\
 +demonstrations& 72.7 & 80.0 \\
\hline
LM-BFF (auto) 	& 71.8 & 79.0 \\
+demonstrations & 72.2 & 80.3 \\
\hline
DART            & 70.8 & 80.8 \\
\hline
DPT             & 54.0 & 79.3 \\
\hline
PromptELECTRA   & 70.1 &  81.0 \\
\hline\hline
DLM-SCS (ours)  & \textbf{76.0} & \textbf{81.6} \\
\hline
\end{tabular}
}
\caption{The mean performance metrics of compared methods. MM-SP denotes the mean performance metric on the 6 sentence-pair tasks, and MM-S denotes the mean performance metric on the 4 single-sentence tasks. }
\label{tab:amcomparison}
\end{table}

\begin{table}
\centering
\scalebox{0.95}{
\begin{tabular}{lc|ccc}
\hline\hline
\textbf{Dataset}  &  \textbf{O.M}  &  \textbf{U.R}  & \textbf{U.M}  & \textbf{D.M}\\
\hline\hline
SNLI (Acc)  & 82.9 & 92.0\% & 85.3 & 41.7 \\
\hline
MNLI (Acc)  & 71.1 & 82.0\% & 75.4 & 41.6 \\
\hline
QNLI (Acc)  & 78.5 & 86.9\% & 81.4 & 37.2 \\
\hline
RTE (Acc)   & 73.3 & 67.5\% & 82.9 & 56.7 \\
\hline
MRPC (F1)   & 79.4 & 83.3\% & 81.6 & 33.9 \\
\hline
QQP (F1)    & 71.4 & 87.5\% & 75.1 & 28.0 \\
\hline
SST-2 (Acc) & 93.3 & 99.0\% & 94.8 & 30.0 \\
\hline
SST-5 (Acc) & 53.9 & 74.3\% & 56.4 & 44.1 \\
\hline
MR (Acc)    & 90.8 & 96.9\% & 91.7 & 34.9 \\
\hline
CR (Acc)    & 91.3 & 93.3\% & 93.5 & 59.3 \\
\hline
\end{tabular}
}
\caption{Performance analysis of unanimous versus disagreed test examples. The column O.M denotes the overall performance metric on the whole test dataset, the column U.R denotes the ratio of unanimous examples in the test dataset, U.M denotes the performance metric on the unanimous test examples, while D.M denotes the performance metric on the disagreed test examples.}
\label{tab:unanimous}
\end{table}
To provide an intuitive understanding of the overall performance lifting extent achieved by our DLM-SCS method, Table~\ref{tab:amcomparison} reports the mean performance metrics averaged on the sentence-pair tasks and the single-sentence tasks. It can be observed that:
\vspace{-2mm}
\begin{itemize}
\setlength\itemsep{0em}
\item On the sentence-pair tasks, a substantial performance lifting is observed. Our \textbf{DLM-SCS} model achieves the highest mean metric score 76.0, superseding the score 72.7 of the runner-up \textbf{LM-BFF(man)+demonstrations} by a large margin. 
\item On the single-sentence tasks, the \textbf{DLM-SCS} model also achieves the highest mean metric 81.6, compared with 81.0 of the runner-up \textbf{PromptELECTRA}.
\end{itemize}

The extent of performance lifting on sentence-pair tasks is much larger than that on single-sentence tasks. 
One potential reason is that it aggregates three kinds of evidence (from the first sentence, the second sentence, and the label word) for sentence-pair tasks. In addition, \textbf{+demonstrations} can improve the performance of \textbf{LM-BFF(man)} and \textbf{LM-BFF(auto)}, because the demonstrations provide more information in the context. As a conclusion, \textbf{Prerequisite 3} of \textit{Evidence Abundance} is highly valuable in few-shot setting.

%\begin{table*}
%\centering
%\small
%\begin{tabular}{lcccccccccc}
%\hline
%Model			& \textbf{SNLI} 	& \textbf{MNLI} & \textbf{QNLI} & \textbf{RTE} & \textbf{MRPC} & \textbf{QQP} & SST-2 & SST-5  & MR  & CR \\
%\hline
%LM-BFF   & 0.73  & 0.688  &  &   & 0.772   &   & & & & \\
%\hline
%LM-BFF (man) 	& 77.2 (3.7) & 68.0 (5.1) & 69.3 (9.9) & 69.1 (3.6) & 74.5 (5.3) &  65.5 (5.3) & & & & \\
%\hline
%LM-BFF (auto) 	& 77.1 (2.1) & 71.2 (2.6) & 69.5 (4.2) & 73.9 (2.2) & 76.2 (2.3) & 67.0 (3.0) & & & & \\
%\hline
%P-Tuning        & 72.3 (3.0) & 61.5 (2.1) & 64.3 (2.8) & -          & 74.5 (7.6) & 65.6 (3.0) &&&&\\  
%\hline
%DART            & 75.8 (3.0) & 67.5 (2.6) & 66.7 (3.7) & -          & 78.3 (4.5) & 67.8 (3.2) &&&& \\
%\hline
%DLM             &  0.777     &  -         & 0.665      & 0.711      & 0.675      &    &&&&\\
%\hline\hline
%Our model       & 82.2 (1.5) & 71.0 (2.0) & 77.0 (2.4) & 75.0 (4.6) & 78.3 (4.9) & 72.2 (1.4) & 93.6 (0.6) & 51.5 (2.0) & 90.2 (0.7) & 91.0 (1.4) \\\hline
%\end{tabular}
%\end{table*}

\begin{figure*}
\centering
\begin{subfigure}{0.195\textwidth}
\includegraphics[width=\textwidth]{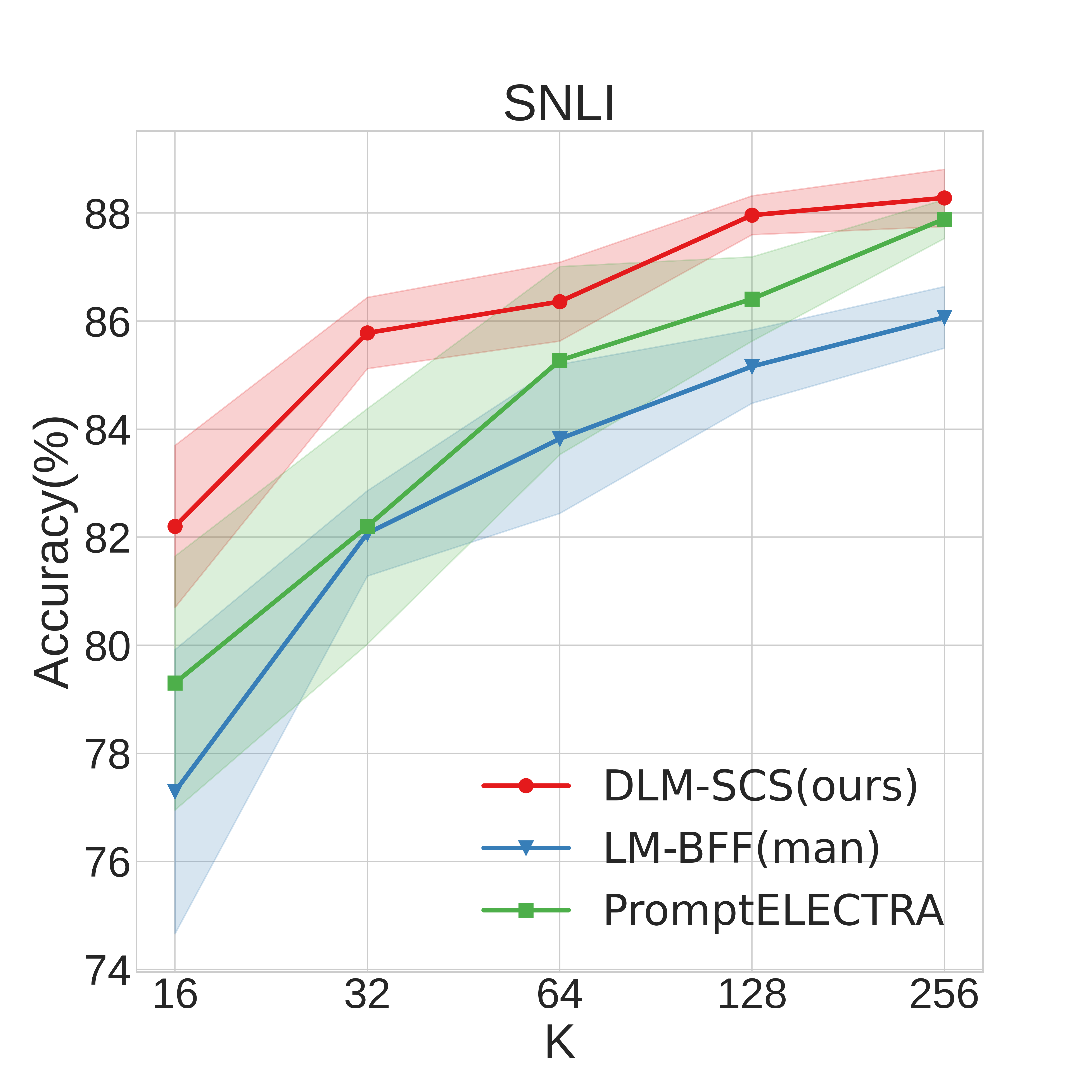}
\end{subfigure}
\begin{subfigure}{0.195\textwidth}
\includegraphics[width=\textwidth]{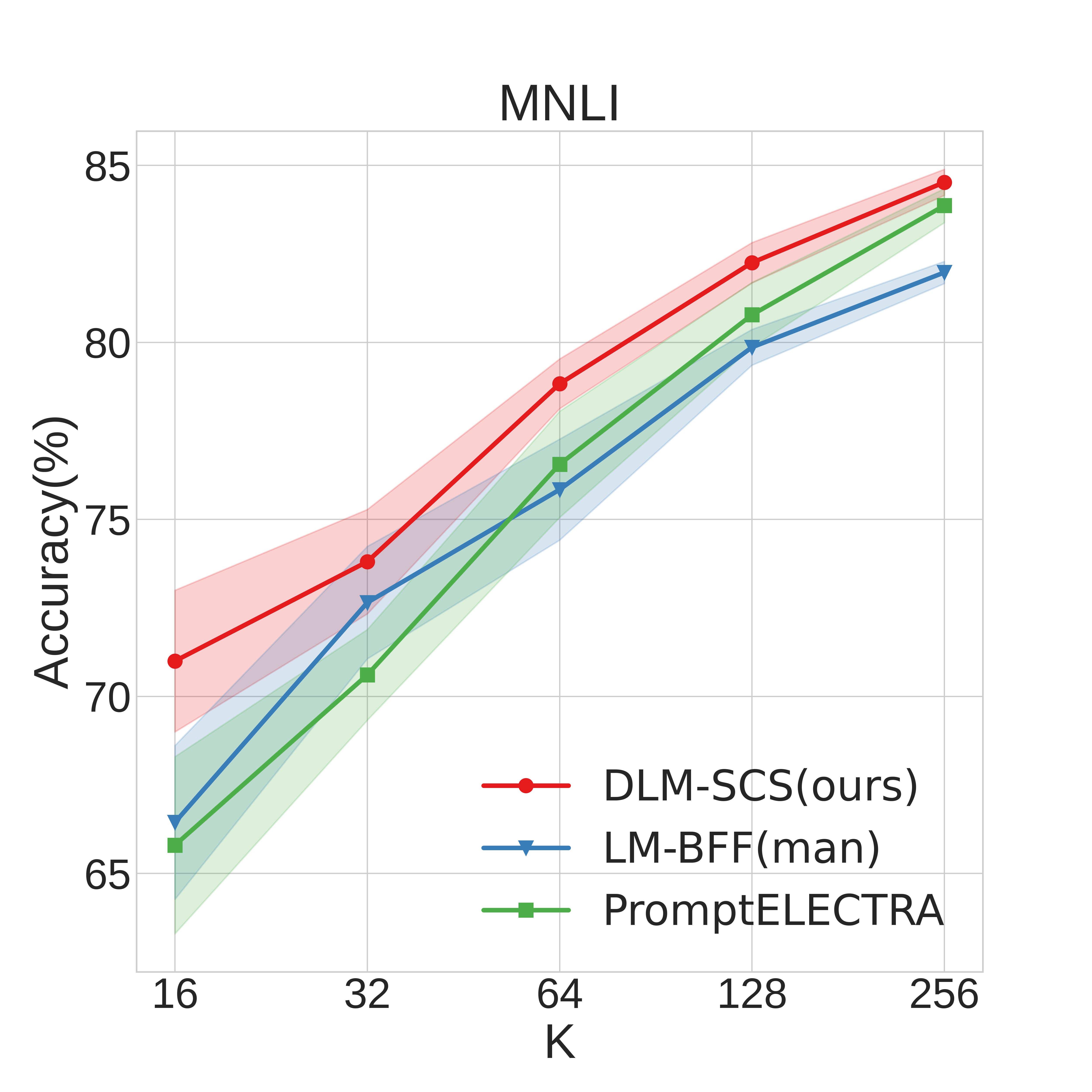}
\end{subfigure}
\begin{subfigure}{0.195\textwidth}
\includegraphics[width=\textwidth]{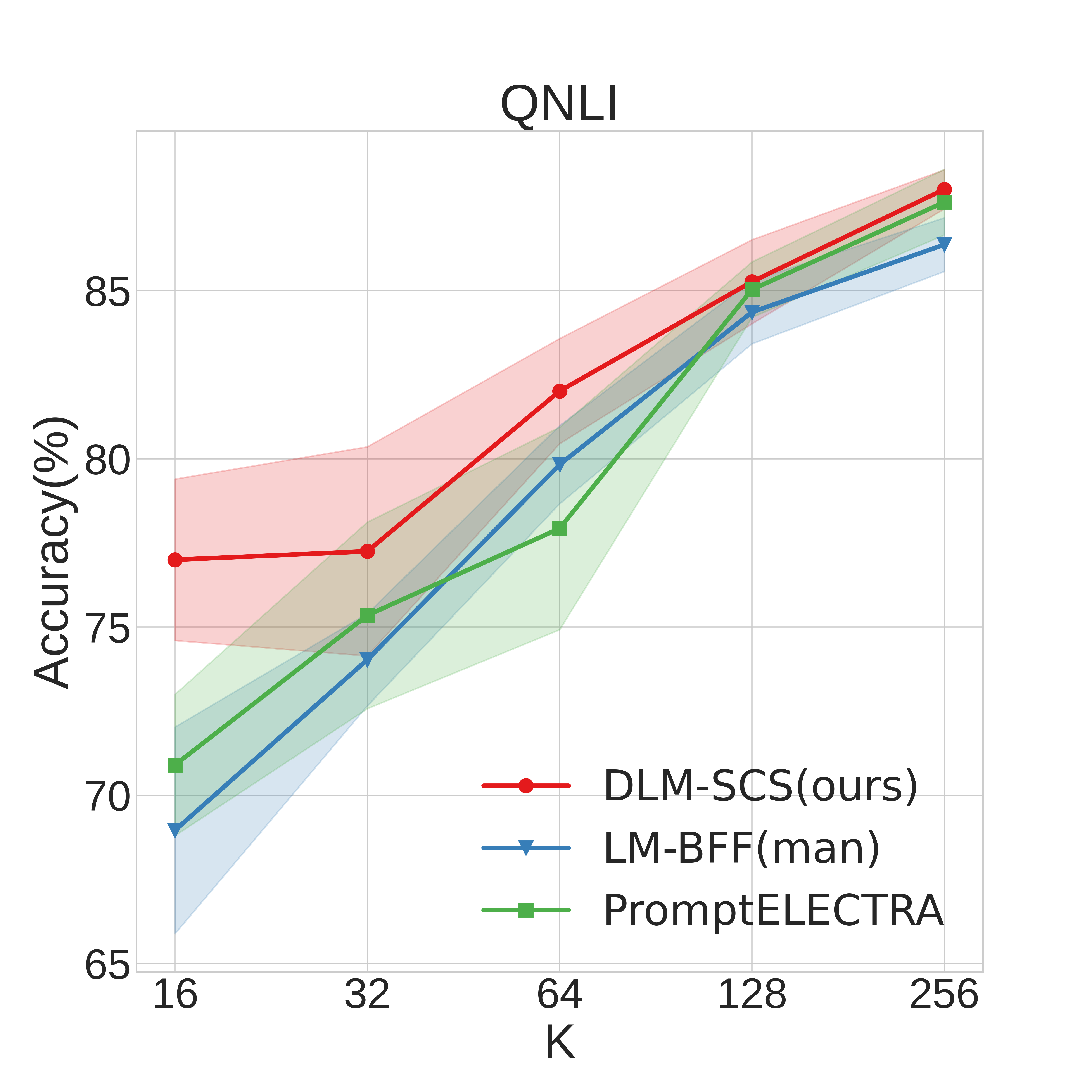}
\end{subfigure}
\begin{subfigure}{0.195\textwidth}
\includegraphics[width=\textwidth]{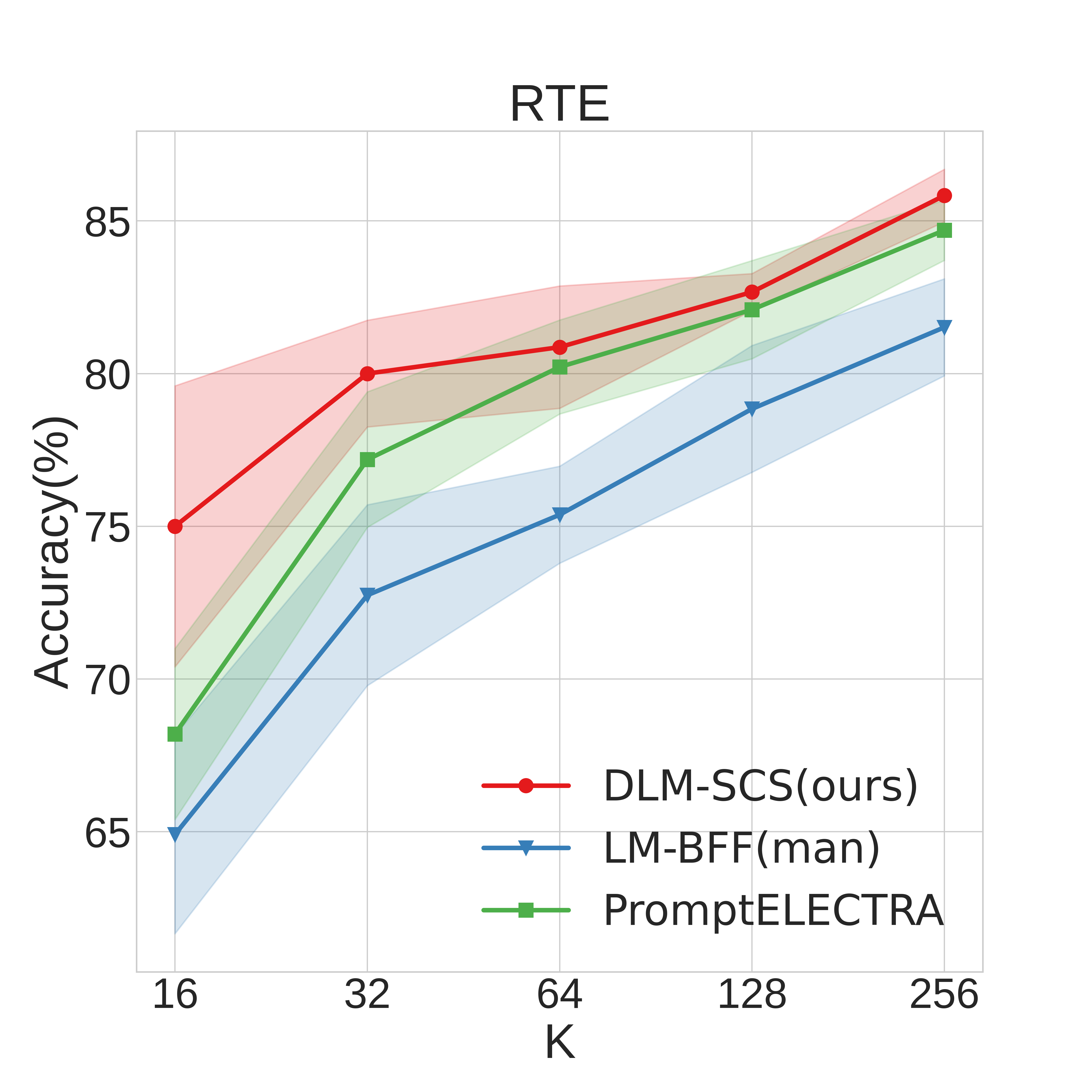}
\end{subfigure}
\begin{subfigure}{0.195\textwidth}
\includegraphics[width=\textwidth]{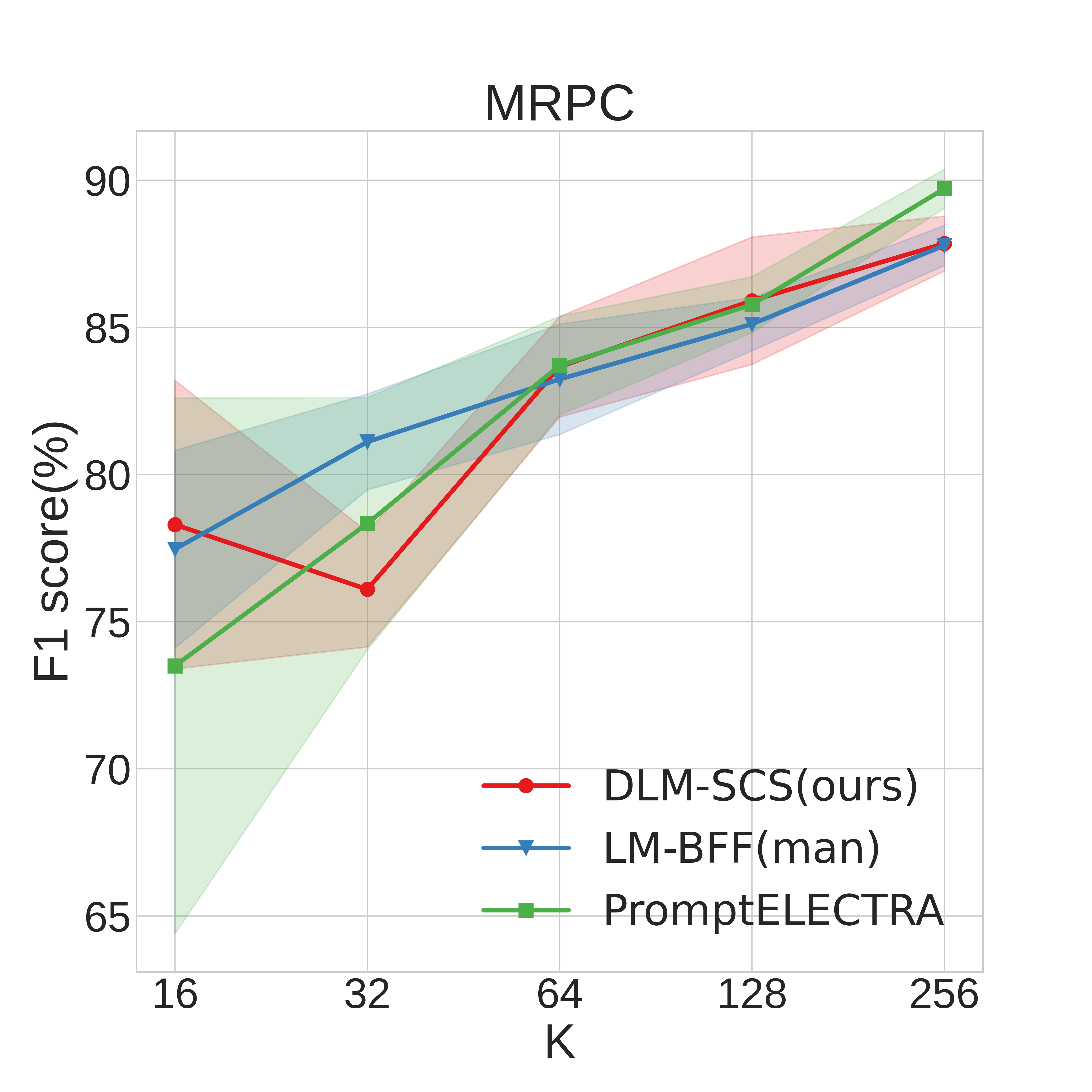}
\end{subfigure}
\begin{subfigure}{0.195\textwidth}
\includegraphics[width=\textwidth]{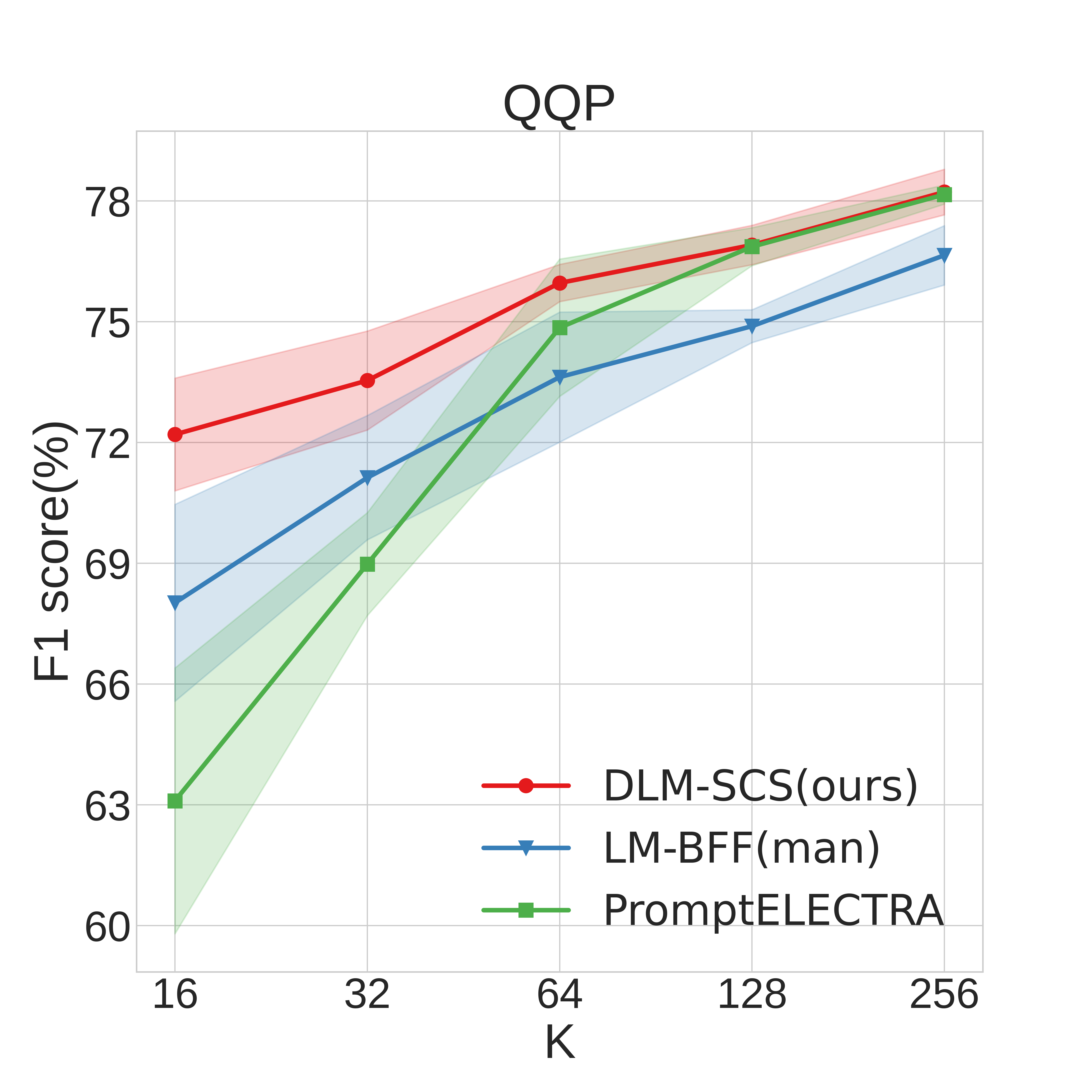}
\end{subfigure}
\begin{subfigure}{0.195\textwidth}
\includegraphics[width=\textwidth]{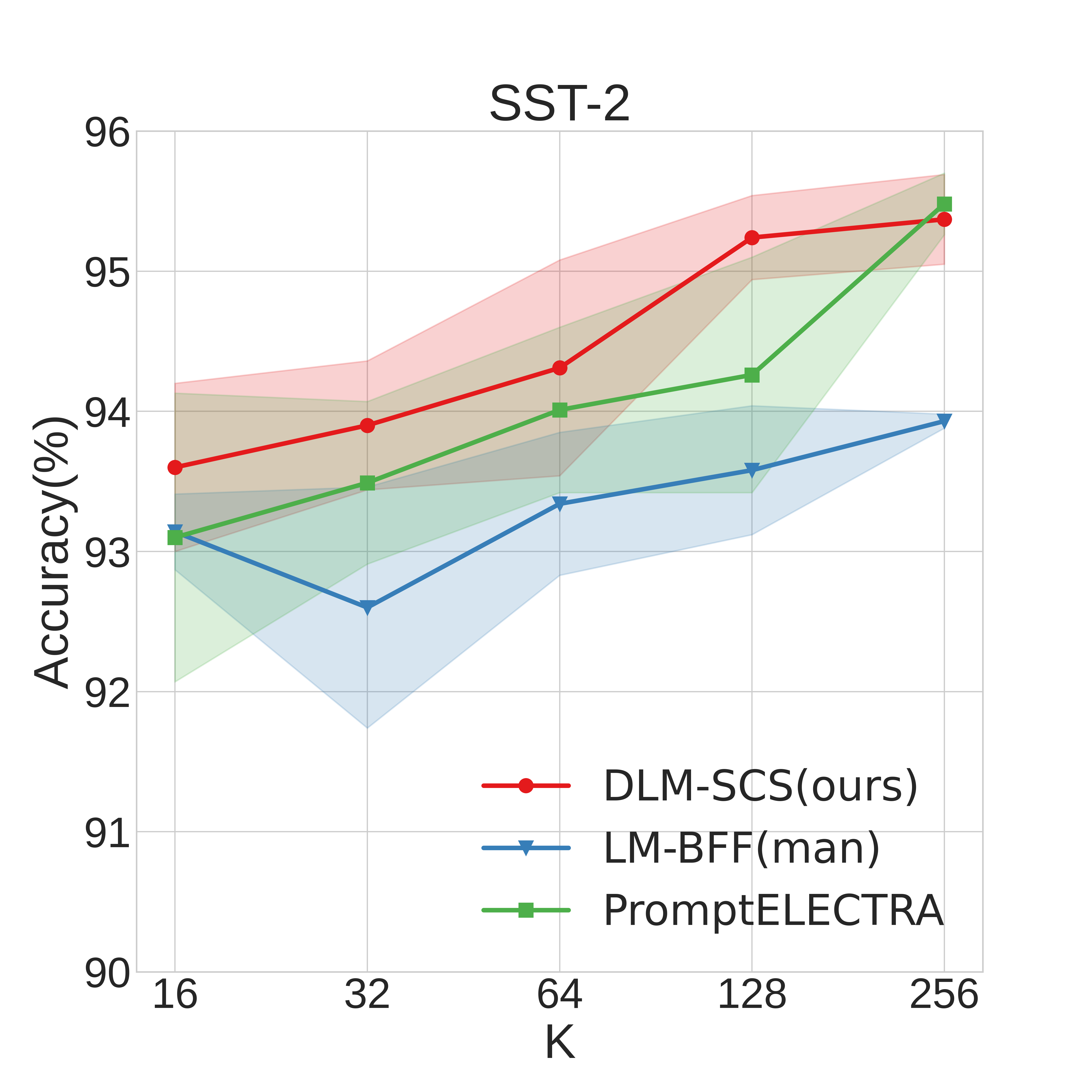}
\end{subfigure}
\begin{subfigure}{0.195\textwidth}
\includegraphics[width=\textwidth]{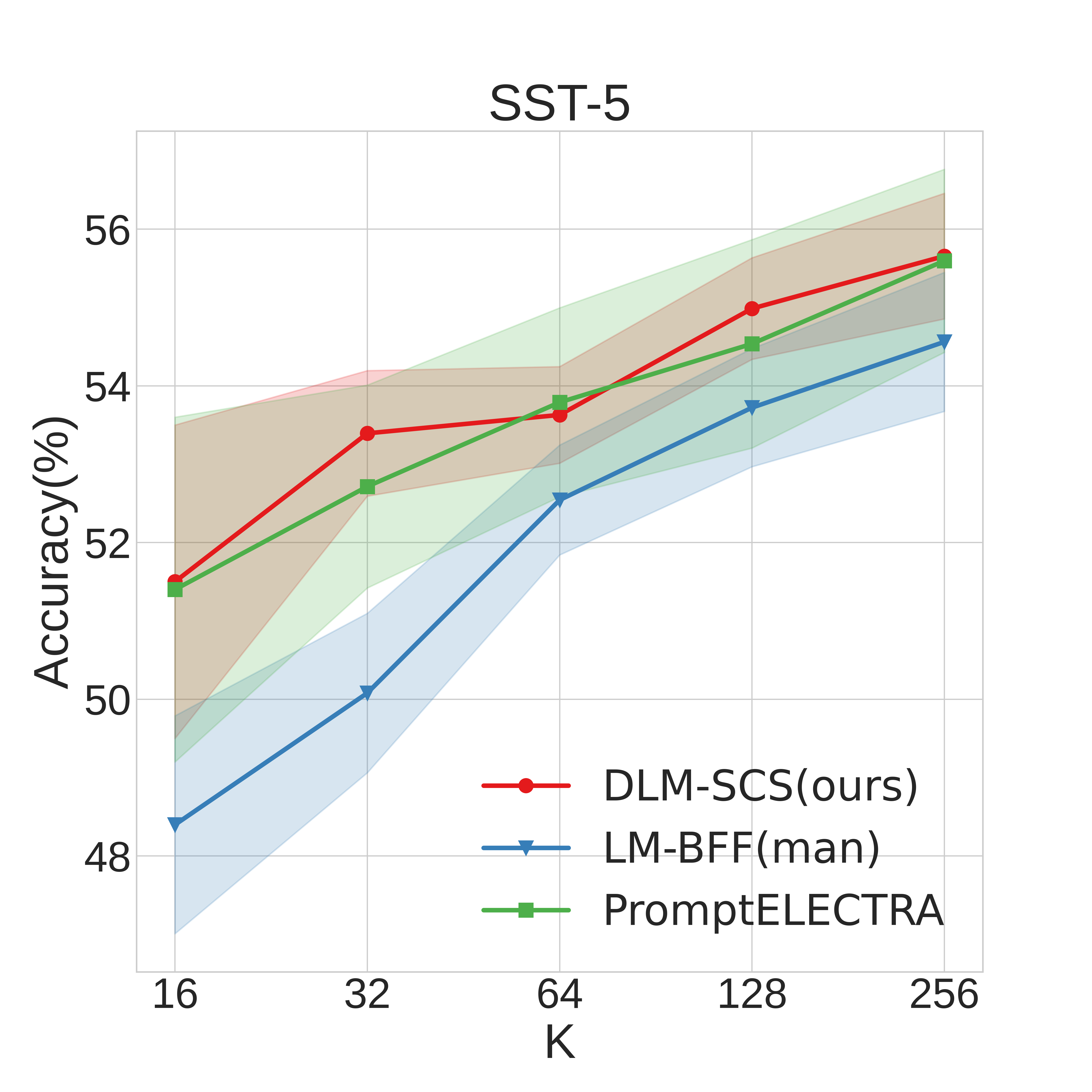}
\end{subfigure}
\begin{subfigure}{0.195\textwidth}
\includegraphics[width=\textwidth]{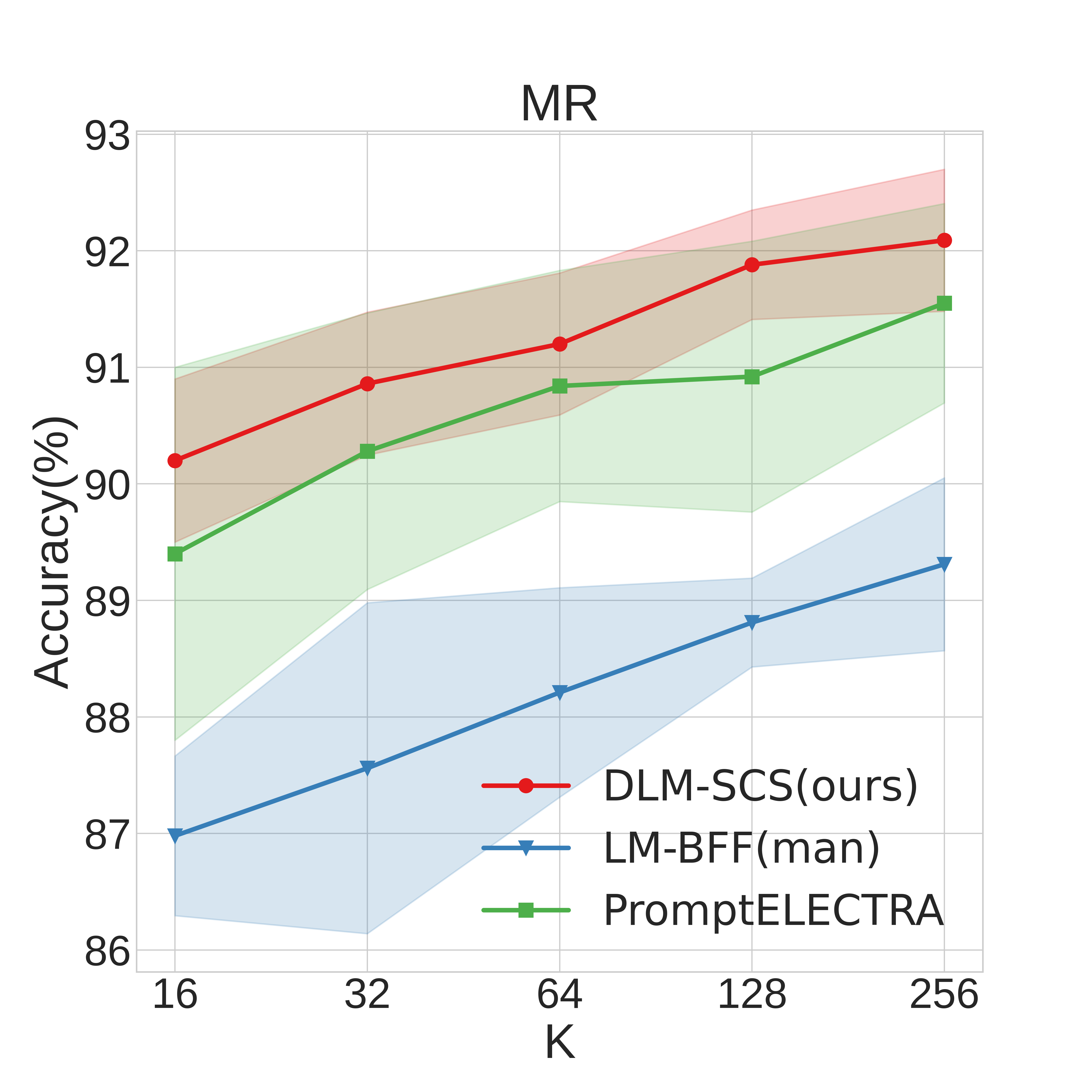}
\end{subfigure}
\begin{subfigure}{0.195\textwidth}
\includegraphics[width=\textwidth]{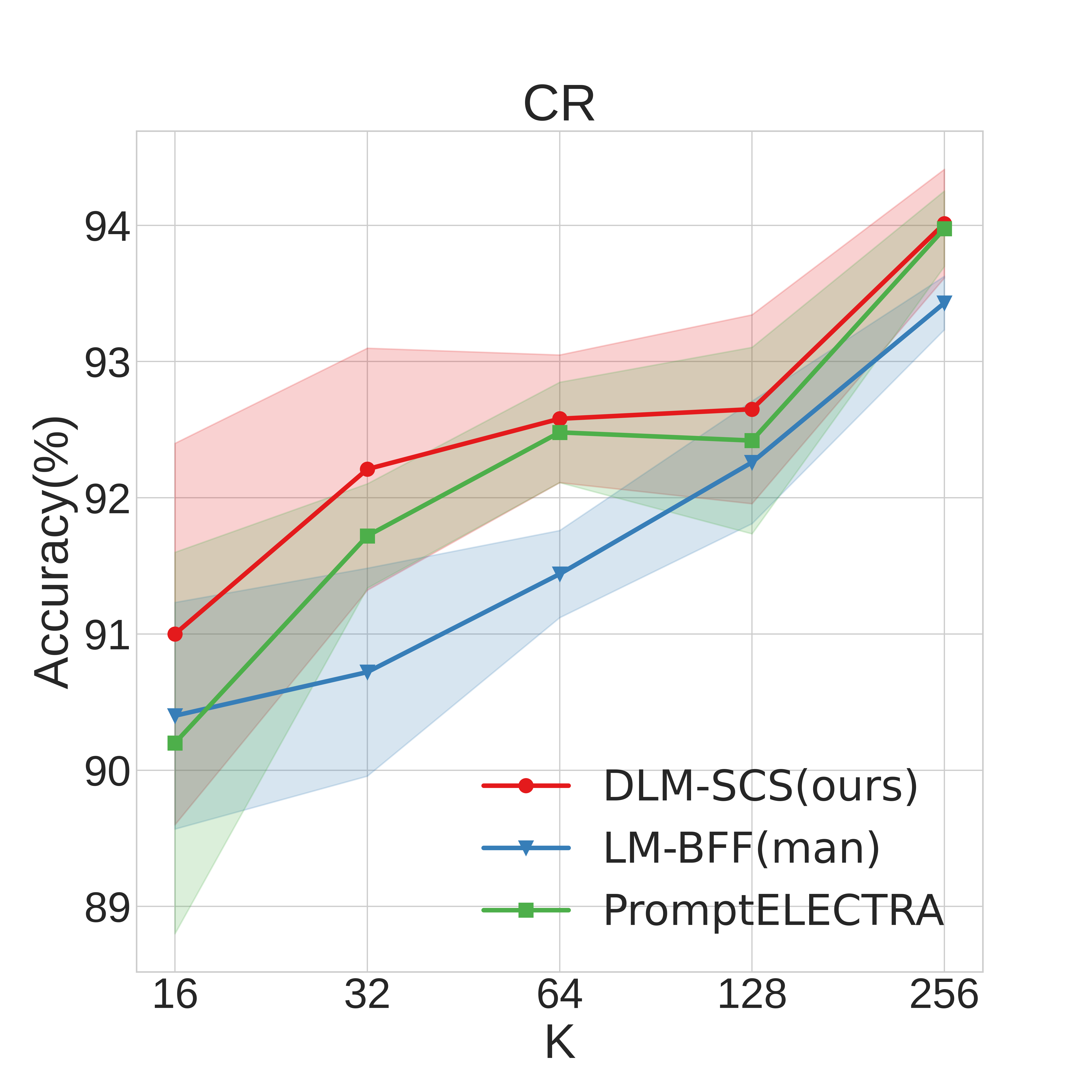}
\end{subfigure}
\caption{Performance comparison of DLM-SCS, LM-BFF (man) and PromptELECTRA with different numbers of training examples. The shadow area denotes the standard deviation of the performance metric.}
\label{fig:varyingK}
\end{figure*}

\subsection{Ablation Study}

Two main techniques have played important roles in the DLM-SCS model: the first is to integrate the evidences from multiple components (or parts) of the prompt, and the second is to weigh the token in each prompt part with IDF values. Table~\ref{tab:ablation} shows the effects on the model performance by removing these two techniques. It can be seen that either of the two techniques is indispensable, because the removal of any technique will lead to substantial reduction of the model performance.

\subsection{Reject Option: Unanimous vs. Disagreed Examples}

The DLM-SCS model for text classification can be thought of as a weighted average of $m$ semantic consistency scorers (refer to Equation~\ref{eq:SC}). For sentence classification task, there are two component scorers ($m=2$), one for the label words and the other for the input sentence. For sentence-pair classification task, there are three component scorers ($m=3$), one for the label words, one for the first sentence, and the other for the second sentence. Therefore, the test examples can be splitted into two parts: an example is called \textit{unanimous} if all component scorers make the same decision for it; otherwise, it is a \textit{disagreed} example, 

\textit{Reject option} means that a classifier refuses to make a decision of a test example if the decision is thought of not sufficiently reliable, which is an important technique to improve the reliability of decision making. Traditional technique for reject option relies on whether the predictive probability (or predictive confidence) is higher than a pre-set threshold, which can be called \textit{quantitative reject option}. However, in few-shot learning scenario, it is almost impossible to obtained reliable predictive probabilities. Instead, DLM-SCS can adopt the following \textit{qualitative reject option} technique: it refuses to make prediction for a disagreed example.

From Table~\ref{tab:unanimous}, it can be seen that the proportion of unanimous examples is high on the whole (larger than 80\% on 8 datasets), which yields to the relatively low reject ratio. In addition, the performance metrics of disagreed examples are much lower than those of unanimous examples, which justifies the refusal of making decision for these disagreed examples.

%\subsection{Ensembling Multiple Prompt Templates}

\subsection{Varying the number of training examples}

Figure~\ref{fig:varyingK} compares the performance \textbf{DLM-SCS} with two competitors \textbf{LM-BFF (man)} and \textbf{PromptELECTRA} as the number of training examples ($K$) increases from 16 to 256. It can be observed that:
\vspace{-2mm}
\begin{itemize}
\setlength\itemsep{0.1em}
\item On 9 datasets (except MRPC), \textbf{DLM-SCS} consistently outperforms \textbf{LM-BFF (man)} when $K$ varies from 16 to 256. 
\item On 7 datasets (except MRPC, SST-2 and SST-5), \textbf{DLM-SCS} consistently outperforms \textbf{PromptELECTRA} for all $K$ values. On SST-2 and SST-5, \textbf{DLM-SCS} wins \textbf{PromptELECTRA} for most $K$ values (4 out of 5).
\end{itemize}

\section{Conclusion and Future Work}

In this paper, we present DLM-SCS, a simple framework for finetuning discriminative language model using only a few examples, where the discriminative language model is used as a semantic consistency scorer of discriminative prompts. 

At its current state, the DLM-SCS method simply works with manual and discrete prompt templates which may be suboptimal. It would be interesting to investigate how to adapt the techniques of automatically prompt generation~\cite{gao2021} and/or differentiable prompting~\cite{zhang2022} to our discriminative framework.

\section*{Limitations}
The main limitation of the proposed DLM-SCS method is that it in the current version cannot be applied to the classification task of a large number of classes, because the GPU memory requirement is linear to the number of classes during training phrase where we have to construct the discriminative prompts of all classes and feed them into GPU. The PromptELECTRA method recently proposed in \cite{xia2022} also suffers from this limitation. How to improve its efficiency and scalability to tasks with many classes would be an interesting work in the future.

In addition, this method works on the basis of a pretrained discriminative language model and is not applicable to the languages without available pretrained DLMs. 

\bibliographystyle{abbrvnat}
\bibliography{discriminativeprompt}

\begin{thebibliography}{19}
\expandafter\ifx\csname natexlab\endcsname\relax\def\natexlab#1{#1}\fi

\bibitem[{Brown et~al.(2020)Brown, Mann, Ryder, Subbiah, Kaplan, Dhariwal,
  Neelakantan, Shyam, Sastry, Askell, Agarwal, Herbert{-}Voss, Krueger,
  Henighan, Child, Ramesh, Ziegler, Wu, Winter, Hesse, Chen, Sigler, Litwin,
  Gray, Chess, Clark, Berner, McCandlish, Radford, Sutskever, and
  Amodei}]{brown2020}
Tom~B. Brown, Benjamin Mann, Nick Ryder, Melanie Subbiah, Jared Kaplan,
  Prafulla Dhariwal, Arvind Neelakantan, Pranav Shyam, Girish Sastry, Amanda
  Askell, Sandhini Agarwal, Ariel Herbert{-}Voss, Gretchen Krueger, Tom
  Henighan, Rewon Child, Aditya Ramesh, Daniel~M. Ziegler, Jeffrey Wu, Clemens
  Winter, Christopher Hesse, Mark Chen, Eric Sigler, Mateusz Litwin, Scott
  Gray, Benjamin Chess, Jack Clark, Christopher Berner, Sam McCandlish, Alec
  Radford, Ilya Sutskever, and Dario Amodei. 2020.
\newblock \href
  {https://proceedings.neurips.cc/paper/2020/hash/1457c0d6bfcb4967418bfb8ac142f64a-Abstract.html}
  {Language models are few-shot learners}.
\newblock In \emph{Advances in Neural Information Processing Systems 33: Annual
  Conference on Neural Information Processing Systems 2020, NeurIPS 2020,
  December 6-12, 2020, virtual}.

\bibitem[{Clark et~al.(2020)Clark, Luong, Le, and Manning}]{clark2020}
Kevin Clark, Minh{-}Thang Luong, Quoc~V. Le, and Christopher~D. Manning. 2020.
\newblock \href {https://openreview.net/forum?id=r1xMH1BtvB} {{ELECTRA:}
  pre-training text encoders as discriminators rather than generators}.
\newblock In \emph{8th International Conference on Learning Representations,
  {ICLR} 2020, Addis Ababa, Ethiopia, April 26-30, 2020}. OpenReview.net.

\bibitem[{Devlin et~al.(2019)Devlin, Chang, Lee, and Toutanova}]{devlin2019}
Jacob Devlin, Ming-Wei Chang, Kenton Lee, and Kristina Toutanova. 2019.
\newblock \href {https://doi.org/10.18653/v1/N19-1423} {{BERT}: Pre-training of
  deep bidirectional transformers for language understanding}.
\newblock In \emph{Proceedings of the 2019 Conference of the North {A}merican
  Chapter of the Association for Computational Linguistics: Human Language
  Technologies, Volume 1 (Long and Short Papers)}, pages 4171--4186,
  Minneapolis, Minnesota. Association for Computational Linguistics.

\bibitem[{Dong et~al.(2019)Dong, Yang, Wang, Wei, Liu, Wang, Gao, Zhou, and
  Hon}]{dong2019}
Li~Dong, Nan Yang, Wenhui Wang, Furu Wei, Xiaodong Liu, Yu~Wang, Jianfeng Gao,
  Ming Zhou, and Hsiao{-}Wuen Hon. 2019.
\newblock \href
  {https://proceedings.neurips.cc/paper/2019/hash/c20bb2d9a50d5ac1f713f8b34d9aac5a-Abstract.html}
  {Unified language model pre-training for natural language understanding and
  generation}.
\newblock In \emph{Advances in Neural Information Processing Systems 32: Annual
  Conference on Neural Information Processing Systems 2019, NeurIPS 2019,
  December 8-14, 2019, Vancouver, BC, Canada}, pages 13042--13054.

\bibitem[{Gao et~al.(2021)Gao, Fisch, and Chen}]{gao2021}
Tianyu Gao, Adam Fisch, and Danqi Chen. 2021.
\newblock \href {https://doi.org/10.18653/v1/2021.acl-long.295} {Making
  pre-trained language models better few-shot learners}.
\newblock In \emph{Proceedings of the 59th Annual Meeting of the Association
  for Computational Linguistics and the 11th International Joint Conference on
  Natural Language Processing, {ACL/IJCNLP} 2021, (Volume 1: Long Papers),
  Virtual Event, August 1-6, 2021}, pages 3816--3830. Association for
  Computational Linguistics.

\bibitem[{Liu et~al.(2021)Liu, Yuan, Fu, Jiang, Hayashi, and Neubig}]{liu2021}
Pengfei Liu, Weizhe Yuan, Jinlan Fu, Zhengbao Jiang, Hiroaki Hayashi, and
  Graham Neubig. 2021.
\newblock \href {http://arxiv.org/abs/2107.13586} {Pre-train, prompt, and
  predict: {A} systematic survey of prompting methods in natural language
  processing}.
\newblock \emph{CoRR}, abs/2107.13586.

\bibitem[{Liu et~al.(2019)Liu, Ott, Goyal, Du, Joshi, Chen, Levy, Lewis,
  Zettlemoyer, and Stoyanov}]{liu2019}
Yinhan Liu, Myle Ott, Naman Goyal, Jingfei Du, Mandar Joshi, Danqi Chen, Omer
  Levy, Mike Lewis, Luke Zettlemoyer, and Veselin Stoyanov. 2019.
\newblock Roberta: A robustly optimized bert pretraining approach.
\newblock \emph{arXiv preprint arXiv:1907.11692}.

\bibitem[{Park et~al.(2022)Park, Jeon, Kim, Kang, and Na}]{park2022}
Eunhwan Park, Dong~Hyeon Jeon, Seonhoon Kim, Inho Kang, and Seung{-}Hoon Na.
  2022.
\newblock \href {https://aclanthology.org/2022.acl-short.34} {{LM-BFF-MS:}
  improving few-shot fine-tuning of language models based on multiple soft
  demonstration memory}.
\newblock In \emph{Proceedings of the 60th Annual Meeting of the Association
  for Computational Linguistics (Volume 2: Short Papers), {ACL} 2022, Dublin,
  Ireland, May 22-27, 2022}, pages 310--317. Association for Computational
  Linguistics.

\bibitem[{Radford et~al.(2018)Radford, Narasimhan, Salimans, Sutskever
  et~al.}]{radford2018}
Alec Radford, Karthik Narasimhan, Tim Salimans, Ilya Sutskever, et~al. 2018.
\newblock Improving language understanding by generative pre-training.
\newblock \emph{OpenAI blog}.

\bibitem[{Radford et~al.(2019)Radford, Wu, Child, Luan, Amodei, Sutskever
  et~al.}]{radford2019}
Alec Radford, Jeffrey Wu, Rewon Child, David Luan, Dario Amodei, Ilya
  Sutskever, et~al. 2019.
\newblock Language models are unsupervised multitask learners.
\newblock \emph{OpenAI blog}, 1(8):9.

\bibitem[{Raffel et~al.(2020)Raffel, Shazeer, Roberts, Lee, Narang, Matena,
  Zhou, Li, and Liu}]{raffel2020}
Colin Raffel, Noam Shazeer, Adam Roberts, Katherine Lee, Sharan Narang, Michael
  Matena, Yanqi Zhou, Wei Li, and Peter~J. Liu. 2020.
\newblock \href {http://jmlr.org/papers/v21/20-074.html} {Exploring the limits
  of transfer learning with a unified text-to-text transformer}.
\newblock \emph{J. Mach. Learn. Res.}, 21:140:1--140:67.

\bibitem[{Schick et~al.(2020)Schick, Schmid, and Sch{\"{u}}tze}]{schick2020}
Timo Schick, Helmut Schmid, and Hinrich Sch{\"{u}}tze. 2020.
\newblock \href {https://doi.org/10.18653/v1/2020.coling-main.488}
  {Automatically identifying words that can serve as labels for few-shot text
  classification}.
\newblock In \emph{Proceedings of the 28th International Conference on
  Computational Linguistics, {COLING} 2020, Barcelona, Spain (Online), December
  8-13, 2020}, pages 5569--5578. International Committee on Computational
  Linguistics.

\bibitem[{Schick and Sch{\"{u}}tze(2021{\natexlab{a}})}]{schick2021}
Timo Schick and Hinrich Sch{\"{u}}tze. 2021{\natexlab{a}}.
\newblock \href {https://doi.org/10.18653/v1/2021.eacl-main.20} {Exploiting
  cloze-questions for few-shot text classification and natural language
  inference}.
\newblock In \emph{Proceedings of the 16th Conference of the European Chapter
  of the Association for Computational Linguistics: Main Volume, {EACL} 2021,
  Online, April 19 - 23, 2021}, pages 255--269. Association for Computational
  Linguistics.

\bibitem[{Schick and Sch{\"{u}}tze(2021{\natexlab{b}})}]{schick2021b}
Timo Schick and Hinrich Sch{\"{u}}tze. 2021{\natexlab{b}}.
\newblock \href {https://doi.org/10.18653/v1/2021.naacl-main.185} {It's not
  just size that matters: Small language models are also few-shot learners}.
\newblock In \emph{Proceedings of the 2021 Conference of the North American
  Chapter of the Association for Computational Linguistics: Human Language
  Technologies, {NAACL-HLT} 2021, Online, June 6-11, 2021}, pages 2339--2352.
  Association for Computational Linguistics.

\bibitem[{Sun et~al.(2021)Sun, Wang, Feng, Ding, Pang, Shang, Liu, Chen, Zhao,
  Lu, Liu, Wu, Gong, Liang, Shang, Sun, Liu, Ouyang, Yu, Tian, Wu, and
  Wang}]{sun2021}
Yu~Sun, Shuohuan Wang, Shikun Feng, Siyu Ding, Chao Pang, Junyuan Shang,
  Jiaxiang Liu, Xuyi Chen, Yanbin Zhao, Yuxiang Lu, Weixin Liu, Zhihua Wu,
  Weibao Gong, Jianzhong Liang, Zhizhou Shang, Peng Sun, Wei Liu, Xuan Ouyang,
  Dianhai Yu, Hao Tian, Hua Wu, and Haifeng Wang. 2021.
\newblock \href {http://arxiv.org/abs/2107.02137} {{ERNIE} 3.0: Large-scale
  knowledge enhanced pre-training for language understanding and generation}.
\newblock \emph{CoRR}, abs/2107.02137.

\bibitem[{Xia et~al.(2022)Xia, Artetxe, Du, Chen, and Stoyanov}]{xia2022}
Mengzhou Xia, Mikel Artetxe, Jingfei Du, Danqi Chen, and Ves Stoyanov. 2022.
\newblock \href {https://doi.org/10.48550/arXiv.2205.15223} {Prompting
  {ELECTRA:} few-shot learning with discriminative pre-trained models}.
\newblock \emph{CoRR}, abs/2205.15223.

\bibitem[{Yao et~al.(2022)Yao, Dong, Zhang, Zhang, Xie, Liu, Lin, Sun, and
  Wang}]{yao2022}
Yuan Yao, Bowen Dong, Ao~Zhang, Zhengyan Zhang, Ruobing Xie, Zhiyuan Liu, Leyu
  Lin, Maosong Sun, and Jianyong Wang. 2022.
\newblock \href {https://aclanthology.org/2022.findings-acl.273} {Prompt tuning
  for discriminative pre-trained language models}.
\newblock In \emph{Findings of the Association for Computational Linguistics:
  {ACL} 2022, Dublin, Ireland, May 22-27, 2022}, pages 3468--3473. Association
  for Computational Linguistics.

\bibitem[{Zhang et~al.(2021)Zhang, Li, Chen, Deng, Bi, Tan, Huang, and
  Chen}]{zhang2021}
Ningyu Zhang, Luoqiu Li, Xiang Chen, Shumin Deng, Zhen Bi, Chuanqi Tan, Fei
  Huang, and Huajun Chen. 2021.
\newblock \href {http://arxiv.org/abs/2108.13161} {Differentiable prompt makes
  pre-trained language models better few-shot learners}.
\newblock \emph{CoRR}, abs/2108.13161.

\bibitem[{Zhang et~al.(2022)Zhang, Li, Chen, Deng, Bi, Tan, Huang, and
  Chen}]{zhang2022}
Ningyu Zhang, Luoqiu Li, Xiang Chen, Shumin Deng, Zhen Bi, Chuanqi Tan, Fei
  Huang, and Huajun Chen. 2022.
\newblock \href {https://openreview.net/forum?id=ek9a0qIafW} {Differentiable
  prompt makes pre-trained language models better few-shot learners}.
\newblock In \emph{International Conference on Learning Representations}.

\end{thebibliography}

\appendix

\section{The Datasets Used}
\label{sec:appdatasets}

This paper uses 10 text classification tasks for experimental evaluation. The 
dataset statistics are listed in Table~\ref{tab:appdatasets}.

\begin{table}[]
\centering
\begin{tabular}{lcrr}
\hline\hline
\textbf{Task}  & $|\mathcal{L}|$    & \#Train   & \#Test         \\ 
\hline\hline
\textbf{SNLI}  &  3  &  549,367  &  9,842    \\ \hline
\textbf{MNLI}  &  3  &  392,702  &  9,815    \\ \hline
\textbf{QNLI}  &  2  &  104,743  &  5,463    \\ \hline
\textbf{RTE}   &  2  &  2,490    &  277      \\ \hline
\textbf{MRPC}  &  2  &  3,668    &  408      \\ \hline
\textbf{QQP}   &  2  &  363,846  &  40,431   \\ \hline
\textbf{SST-2} &  2  &  6,920    &  872      \\ \hline
\textbf{SST-5} &  5  &  8,544    &  2,210    \\ \hline
\textbf{MR}    &  2  &  8,662    &  2,000    \\ \hline
\textbf{CR}    &  2  &  1,775    &  2,000    \\ \hline
\end{tabular}
\caption{Characteristics of the datasets evaluated in this work. $|\mathcal{L}|$ denotes the number classes for the classification task. For few-shot classification setting, the training set $\mathcal{D}_\text{train}$ and the development set $\mathcal{D}_\text{dev}$ are sampled from the original training set, each with $K\times |\mathcal{L}|$ examples. }
\label{tab:appdatasets}
\end{table}

\section{The Manual Templates Used}
\label{sec:apptemplates}

Our work is based on manual templates and label words. The manual templates and label words for each dataset are summarized in Table~\ref{tab:apptemplates}. These templates and
label words are the same as the ones used in \cite{gao2021}.

\begin{table}[]
\centering
\begin{tabular}{lll}
\hline\hline
\textbf{Task}  & \textbf{Template}                                                      & \textbf{Label words}         \\ 
\hline\hline
\textbf{SNLI}  & \textless{}$S_1$\textgreater{}? $v(l)$, \textless{}$S_2$\textgreater{} & Yes/No/Maybe                 \\ \hline
\textbf{MNLI}  & \textless{}$S_1$\textgreater{}? $v(l)$, \textless{}$S_2$\textgreater{} & Yes/No/Maybe                 \\ \hline
\textbf{QNLI}  & \textless{}$S_1$\textgreater{}? $v(l)$, \textless{}$S_2$\textgreater{} & Yes/No                       \\ \hline
\textbf{RTE}   & \textless{}$S_1$\textgreater{}? $v(l)$, \textless{}$S_2$\textgreater{} & Yes/No                       \\ \hline
\textbf{MRPC}  & \textless{}$S_1$\textgreater{}? $v(l)$, \textless{}$S_2$\textgreater{} & Yes/No                       \\ \hline
\textbf{QQP}   & \textless{}$S_1$\textgreater{}. $v(l)$, \textless{}$S_2$\textgreater{} & Yes/No                       \\ \hline
\textbf{SST-2} & \textless{}$S_1$\textgreater It is $v(l)$.                             & terrible/great               \\ \hline
\multirow{2}{*}{\textbf{SST-5}} & \multirow{2}{*}{\textless{}$S_1$\textgreater It is $v(l)$.} & terrible/bad/\\
               &                                                                             & okay/good/great \\   \hline
\textbf{MR}    & \textless{}$S_1$\textgreater It is $v(l)$.                             & terrible/great               \\ \hline
\textbf{CR}    & \textless{}$S_1$\textgreater It is $v(l)$.                             & terrible/great               \\ \hline
\end{tabular}
\caption{The Manual templates and label words used in the experiments.}
\label{tab:apptemplates}
\end{table}
%This is a section in the appendix.

\section{Hyperparameters Used}
\label{sec:hyperparameter}

As described in Section~\ref{sec:optimization}, we use grid search to seek for the optimal hyperparameter configuration based on development set.  Table~\ref{tab:hyperparameter} lists $\lambda_0$ of 5 seeds on all 10 datasets and their average value. Accordingly, $\lambda_1$ (and $\lambda_ 2$) can be calculated based on the value of $\lambda_0$. It can be clearly found from the data that the model prefers to select label word as the discrimination evidence for single sentence classification tasks, while tends to select the input sentences as the determining criterion for sentence pair classification tasks. This can be explained as that in the sentence pair tasks, two input sentences usually contain several synonymous or antisense words which could be vital for discrimination.

\begin{table}[]
\centering
\begin{tabular}{l|ccccc|l}
\hline\hline
\multirow{2}{*}{\textbf{Task}} & \multicolumn{5}{|c|}{Random Seed} & \multirow{2}{*}{\textbf{Avg}} \\\cline{2-6}
  & \textbf{13} & \textbf{21} & \textbf{42} & \textbf{87} & \textbf{100} &  \\ \hline \hline
\textbf{SNLI}  & 0.24        & 0.21        & 0.10         & 0.13        & 0.03         & 0.14         \\
\textbf{MNLI}  & 0.28        & 0.00        & 0.07        & 0.03        & 0.14         & 0.10         \\
\textbf{QNLI}  & 0.00        & 0.00        & 0.00        & 0.03        & 0.00         & 0.01         \\
\textbf{RTE}   & 0.00        & 0.00        & 0.00        & 0.07        & 0.00         & 0.01         \\
\textbf{MRPC}  & 0.89        & 0.00        & 0.41        & 0.21        & 0.00         & 0.30         \\
\textbf{QQP}   & 0.03        & 0.31        & 0.00        & 0.07        & 0.03         & 0.09         \\
\textbf{SST-2} & 1.00        & 1.00        & 1.00        & 0.69        & 0.89         & 0.92         \\
\textbf{SST-5} & 0.73        & 0.51        & 0.73        & 0.41        & 0.34         & 0.54         \\
\textbf{MR}    & 0.24        & 1.00        & 0.89        & 0.69        & 0.10         & 0.58         \\
\textbf{CR}    & 0.21        & 1.00        & 0.62        & 1.00        & 1.00         & 0.77         \\ \hline
\end{tabular}
\caption{The hyperparameter values of $\lambda_0$ chosen by the grid search. }
\label{tab:hyperparameter}
\end{table}

\end{document}